\pdfoutput=1

\documentclass[11pt]{article}

\usepackage[final]{acl}

\usepackage{times}
\usepackage{latexsym}

\usepackage[T1]{fontenc}
\usepackage{fontawesome}

\usepackage[utf8]{inputenc}

\usepackage{microtype}

\usepackage{inconsolata}
\usepackage{multirow}
\usepackage{adjustbox}
\usepackage{linguex}
\usepackage{makecell}
\usepackage[breakable]{tcolorbox}
\usepackage{CJKutf8}
\usepackage{amsmath}

\title{GAOKAO-MM: A Chinese Human-Level Benchmark for \\Multimodal Models Evaluation}

\author{Yi Zong, Xipeng Qiu\footnotemark[1] \\
School of Computer Science, Fudan University \\
Shanghai Collaborative Innovation Center of Intelligent Visual Computing \\
\text{yzong22@m.fudan.edu.cn, xpqiu@fudan.edu.cn}
}

\begin{document}
\maketitle

\renewcommand{\thefootnote}{\fnsymbol{footnote}}
\footnotetext[1]{Corresponding Author.}
\renewcommand{\thefootnote}{\arabic{footnote}}

\begin{abstract}
The Large Vision-Language Models (LVLMs) have demonstrated great abilities in image perception and language understanding. However, existing datasets either focus solely on primary perception abilities and commonsense knowledge, or have a low level of text comprehension difficulty, which are insufficient to reflect the comprehensive capabilities of LVLMs, particularly in terms of Chinese language proficiency. We propose GAOKAO-MM, a multimodal benchmark based on the Chinese College Entrance Examination (GAOKAO), comprising of 8 subjects and 12 types of images, such as diagrams, function graphs, maps and photos. GAOKAO-MM derives from native Chinese context and sets human-level requirements for the model's abilities, including perception, understanding, knowledge and reasoning. We evaluate 10 LVLMs and find that the accuracies of all of them are lower than 50\%, with GPT-4-Vision (48.1\%), Qwen-VL-Plus (41.2\%) and Gemini-Pro-Vision (35.1\%) ranking in the top three positions. The results of our multi-dimension analysis indicate that LVLMs have moderate distance towards Artificial General Intelligence (AGI) and provide insights facilitating the development of multilingual LVLMs.\footnote{The dataset and evaluation code are available through: \url{https://github.com/OpenMOSS/GAOKAO-MM}}

\end{abstract}

\section{Introduction}
Over the past years, large language models (LLMs) (\citealp{brown2020language}; \citealp{OpenAI2023GPT4TR}; \citealp{bubeck2023sparks}; \citealp{anil2023palm}) have achieved great success and have been on par with or surpassed human in certain areas. Inspired by the development of LLMs, large vision-language models (LVLMs) also make much progress on perception, understanding and reasoning, which shows great potential towards Artificial General Intelligence (AGI). However, previous multimodal datasets, including VQA \citep{goyal2017making}, OK-VQA \citep{marino2019okvqa}, TextVQA \citep{singh2019vqa}, focus on a single capability or image type. Consequently, there is a growing need for a more human-level benchmark to evaluate LVLMs on comprehensive tasks. \citeauthor{lu2024mathvista} propose MathVista to evaluate the mathmatical reasoning of foundation models; MME \citep{fu2023mme} measures both perception and cognition abilities of LVLMs; MM-Vet \citep{yu2023mmvet} defines 6 core VL capabilities and examines the 16 integrations of interest.\\


\begin{figure}[!t]
    \centering
    \begin{adjustbox}{center}
    \includegraphics[width=1\linewidth]{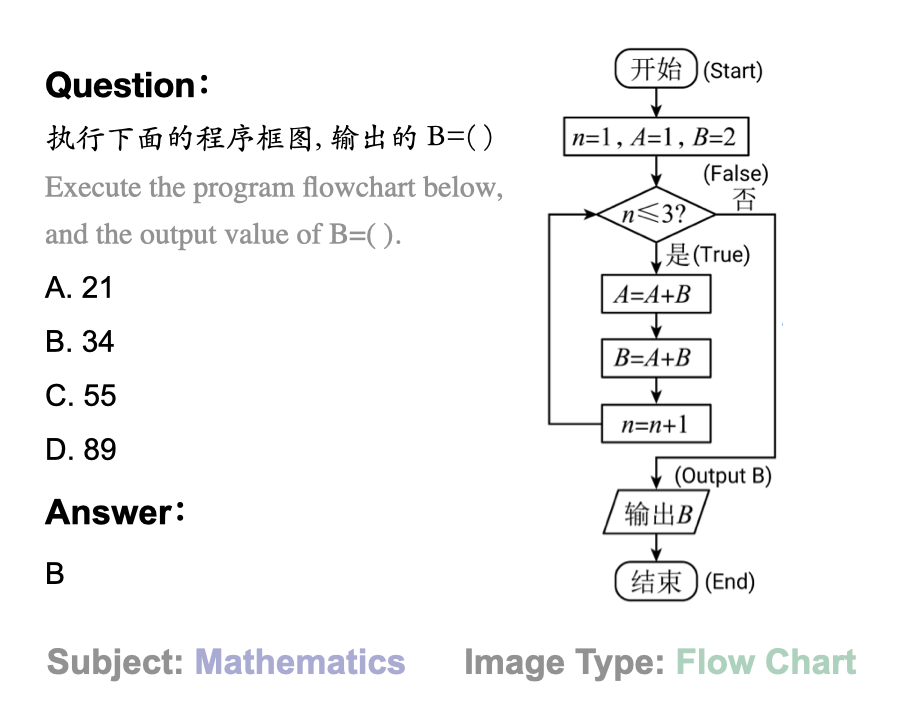}
    \end{adjustbox}
    \caption{\textbf{An example of math question in GAOKAO-MM.} The English translation in the text and images is added by the author for readers' understanding.}
    \label{fig:math data example}
\end{figure}

In this work, we propose GAOKAO-MM, a benchmark derived from Chinese College Entrance Examination (GAOKAO). 
It comprises of 8 subjects, including Chinese, mathematics, physics, chemistry, biology, politics, history and geography, and its average question length is 4 times longer than existing benchmarks. 
GAOKAO-MM covers 12 types of images, such as diagrams, charts, photos and maps. 
LVLMs need to perceive and understand the images and text, fuse their inner knowledge and make inference to get correct answers, which more closely mirrors the way humans perceive the world and make decisions. 
We evaluate 10 LVLMs, including open-source and closed-source models, and the results demonstrate the existing models still have considerable room for improvement.

\section{GAOKAO-MM}
\subsection{Dataset Description}
The GAOKAO, Chinese College Entrance Examination, serves as a pivotal assessment tool for evaluating the scholarly aptitudes of high school students. The GAOKAO is the most popular Chinese standardized test and we regard it as a suitable evaluation on human-level tasks. \\
The GAOKAO-MM comprises of 646 questions and covers 8 subjects, including Chinese, mathematics, physics, chemistry, biology, politics, history and geography. There are 897 images in 12 types, such as diagrams, function graphs, maps, line charts and geometric shapes. The distribution of image types is shown in Appendix \ref{app:statistics}.  \\
The GAOKAO-MM poses a significant challenge for LVLMs: models are supposed to perceive and understand both the image and text inputs, recall their inner knowledge and conduct reasoning to answer the rigorous and comprehensive questions. Thanks to the popularity and authoritative of the GAOKAO, GAOKAO-MM provides a human-level approach to evaluate the capabilities of LVLMs. An example of math question is shown in Figure \ref{fig:math data example}.

\subsection{Data Collection}

\textbf{Data Source}\quad We collect the GAOKAO papers from 2010 to 2023 in PDF format from the Internet.\\
\textbf{Data Collection}\quad We manually extract the multi-choice questions with images using OCR, and the mathematical formulas within the questions were converted into  \LaTeX \, format.\\
\textbf{Data Curation} \quad We select the questions that require the integration of images for accurate answering and filter out duplicate questions. And we perform Lanczos resampling on low-resolution images, ensuring that at least one dimension of the image is increased to no less than 512 pixels. 


\subsection{Comparisons with Existing Benchmarks}
We compare GAOKAO-MM with existing benchmarks which are knowledge-based (\citealp{lu2022iconqa}; \citealp{marino2019okvqa}, \citealp{kembhavi2016diagram}; \citealp{kahou2018figureqa}; \citealp{lu2022learn}; \citealp{yue2023mmmu}; \citealp{liu2023mmbench}). 
And GAOKAO-MM is different from them from three aspects.\\
Firstly, the average question length of GAOKAO-MM is much longer than existing benchmarks, even more than 4 times to the longest one. Previous multimodal benchmarks focus more on the visual perception, while the text input is just an auxiliary to images; GAOKAO-MM puts emphasis both on image perception and text understanding, which simulates the way human receive multimodal information more closely. \\
Secondly, GAOKAO-MM provides detailed explanations of the correct answers annotated by human. And we provide an example of explanation in Appendix \ref{app:example of question and explanations}.\\
Lastly, most Chinese multimodal benchmarks are translated from English benchmarks. However, GAOKAO-MM derives from native Chinese context and contains questions more related to Chinese culture. 

\renewcommand{\arraystretch}{1.5}
\begin{table}[t]\huge
\centering
\resizebox{1\linewidth}{!}{
\begin{tabular}{lcccccc}
\hline
\textbf{Benchmark}& \textbf{Size}&\textbf{Avg. Q. Leng.} &\textbf{Expl.} &\textbf{Question} &\textbf{Image} &\textbf{Lang.}\\ 
\hline
\textbf{IconQA}& 107K & 8.30 & \faTimes & MC+FB &Icon Image& Eng.\\
\textbf{OK-VQA}& 14K &8.09& \faTimes & Open & Photos  & Eng.   \\
\textbf{Ai2D}& 5K &9.78&\faTimes & MC &Diagrams  &  Eng.  \\
\textbf{FigureQA}& $\textgreater$1M &6.07 & \faTimes & BC & 5 Types  &Eng.  \\
\textbf{ScienceQA}& 6K &12.11& \faCheck & MC &  5 Types &Eng.    \\
\textbf{MMMU} & 11.5K &59.33& \faCheck &MC+Open & 30 Types & Eng.\\
\textbf{MM-Bench-CN}& 3K &15.48&\faTimes& MC & Not Mentioned & T. Chi.  \\
\hline
\textbf{GAOKAO-MM} & 0.65K &260.19 & \faCheck & MC & 12 Types & N. Chi. \\
\hline
\end{tabular}
}
\caption{\textbf{The comparison between GAOKAO-MM and other existing benchmarks.} Avg Q. Leng. stands for Average Question Length; FB stands for Fill-in-the-Blank Questions; MC stands for Multiple-Choice Quesions; BC stands for Binary-Choice Questions; T. Chi. stands for translated Chinese; N. Chi. stands for native Chinese.}
\label{tab:Comparison with exsiting benchmarks}
\end{table}
\renewcommand{\arraystretch}{1} 

\renewcommand{\arraystretch}{1}
\begin{table*}[!htbp]
\centering
\setcellgapes{5pt} 
\makegapedcells 
\resizebox{\textwidth}{!}{
\begin{tabular}{l|c|cccccccc} 
\hline
\textbf{Models} & \textbf{Overall} & \textbf{Chinese} & \textbf{Math.} & \textbf{Phys.} & \textbf{Chem.} & \textbf{Biol.} & \textbf{Poli.} & \textbf{Hist.} & \textbf{Geog.}\\ \hline

\textbf{VisCPM-Chat} & 16.8\%  & 0.0\% & 18.8\% & 3.7\% & 22.4\% & 23.8\% & 3.0\% & 32.4\% & 24.9\% \\

\textbf{VisualGLM-6B*} &20.7\% & \textbf{21.9}\% & 16.3\% & 12.1\% & 25.4\% & 28.6\% & 21.2\% & 38.2\% & 24.0\% \\

\textbf{CogVLM-Chat*} &13.2\%  & 3.1\% & 17.5\% & 0.0\% & 16.4\% & 9.5\% & 12.1\% & 20.6\% & 21.3\%  \\

\textbf{LLaVA-1.5-7B*} &17.9\%  & 6.2\% & 12.5\% & 7.2\% & 23.9\% & 14.3\% & 21.2\% & 32.4\% & 24.9\%  \\

\textbf{LLaVA-1.5-13B*} & 18.0\% & 3.1\% & 16.3\% & 10.6\% & 19.4\% & 33.3\% & 12.1\% & 23.5\% & 23.5\% \\ 

\textbf{mPLUG-Owl2} & 24.1\% & 6.2\% & 22.5\% & 12.4\% & 22.4\% & 28.6\% & 48.5\% & 32.4\% & 30.3\% \\ 

\textbf{Qwen-VL-Chat} & 14.2\% & 0.0\% & 17.5\% & 3.2\% & 16.4\% & 14.3\% & 15.2\% & 26.5\% & 19.9\% \\ 

\hline

\textbf{GPT-4V} & \textbf{48.1}\% & 18.8\% & \textbf{45.0}\% & \textbf{26.7}\% & 41.8\% & 38.1\% & \textbf{75.8}\% & 79.4\% & \textbf{62.0}\% \\

\textbf{Qwen-VL-Plus} & 41.2\% & 18.8\% & 33.8\% & 14.4\% & \textbf{44.8}\% & \textbf{42.9}\% & 63.6\% & \textbf{82.4}\% & 55.7\% \\ 

\textbf{Gemini-Pro-Vision} & 35.1\% & 18.8\% & 23.7\% & 17.2\% & 38.8\% & \textbf{42.9}\% & 54.5\% & 55.9\% & 46.6\% \\ 
\hline
\end{tabular}
}
\caption{\textbf{Accuracy of LVLMs on GAOKAO-MM.}The models above the line are open-source models, models below the line are closed-source models. *: Model does not support the simultaneous input of multiple images.}
\label{tab:accuracy}
\end{table*}
\renewcommand{\arraystretch}{1} 

\section{Experiments}
\subsection{Methodology}
We evaluate 10 LVLMs on GAOKAO-MM, including open-source models and closed-source models. Following GAOKAO-Bench \citep{zhang2023evaluating}, we utilize a zero-shot setting strategy to emulate how human partake in examinations. 
We design prompts tailored to different subjects and adopt a robust and rule-based approach to extract answers. 
Models are prompted to output the reasoning process and final answers in an explicit format. 
The examples of prompts are shown in Appendix \ref{app:examples of prompts} and examples of model outputs are shown in Appendix \ref{app:Examples of Model Outputs}.\\ \\
\textbf{Models}\quad We evaluate 7 open-source and 3 closed-source LVLMs on GAOKAO-MM, and deploy the open-source models through their checkpoints from Huggingface and access closed-source models via their API calls.\\ \\
Open-source models include: 
(1) \textit{VisCPM-Chat} \citep{hu2024large}; 
(2) \textit{VisualGLM-6B} \citep{du2022glm}; 
(3) \textit{CogVLM-Chat} \citep{wang2024cogvlm}; 
(4) \textit{LLaVA-1.5} \citep{liu2023improved} We test on its 7B and 13B versions; 
(5) \textit{mPLUG-Owl2} \citep{ye2023mplugowl2}; 
(6) \textit{Qwen-VL-Chat} \citep{bai2023qwenvl}.\\ \\
Closed-source models are: 
(1) \textit{GPT-4V} \citep{OpenAI2023GPT4TR}; 
(2) \textit{Qwen-VL-Plus} \citep{bai2023qwenvl}; 
(3) \textit{Gemini-Pro-Vision} \citep{geminiteam2023gemini}.



\subsection{Results}
\textbf{Overall Performance}\quad The accuracies of all of the LVLMs are lower than 50\%, and GPT-4V ranks the first with an accuracy of 48.1\%. This reflects GAOKAO-MM poses a considerable challenge to the multimodal models due to its high demands on human-level capabilities, including perception, understanding, knowledge and reasoning. \\ \\
\textbf{Performance of Open-Sourced and Closed-Source Models}\quad All of the open-source models reach accuracy rates lower than 25\%, and the gap between the open-source and closed-source models is greater than 11\%. This indicates there is still huge room for improvement for the open-source community. 

\begin{figure}[!htbp]
    \centering
    \begin{adjustbox}{center}
    \includegraphics[width=0.9\linewidth]{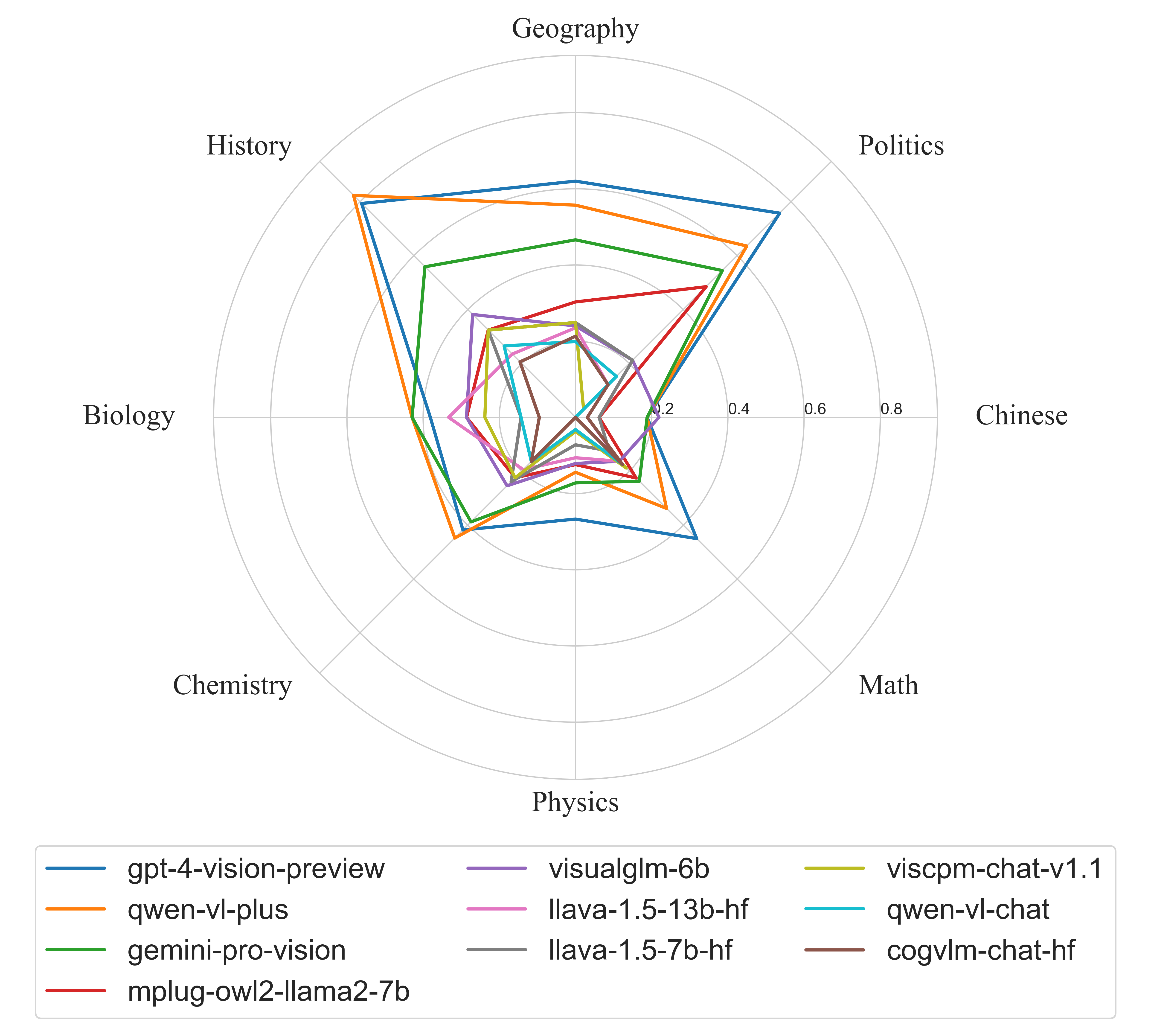}
    \end{adjustbox}
    \caption{Different Performance in Subjects.}
    \label{fig:difference in subject}
\end{figure}

\begin{figure}[!htbp]
    \centering
    \begin{adjustbox}{center}
    \includegraphics[width=0.9\linewidth]{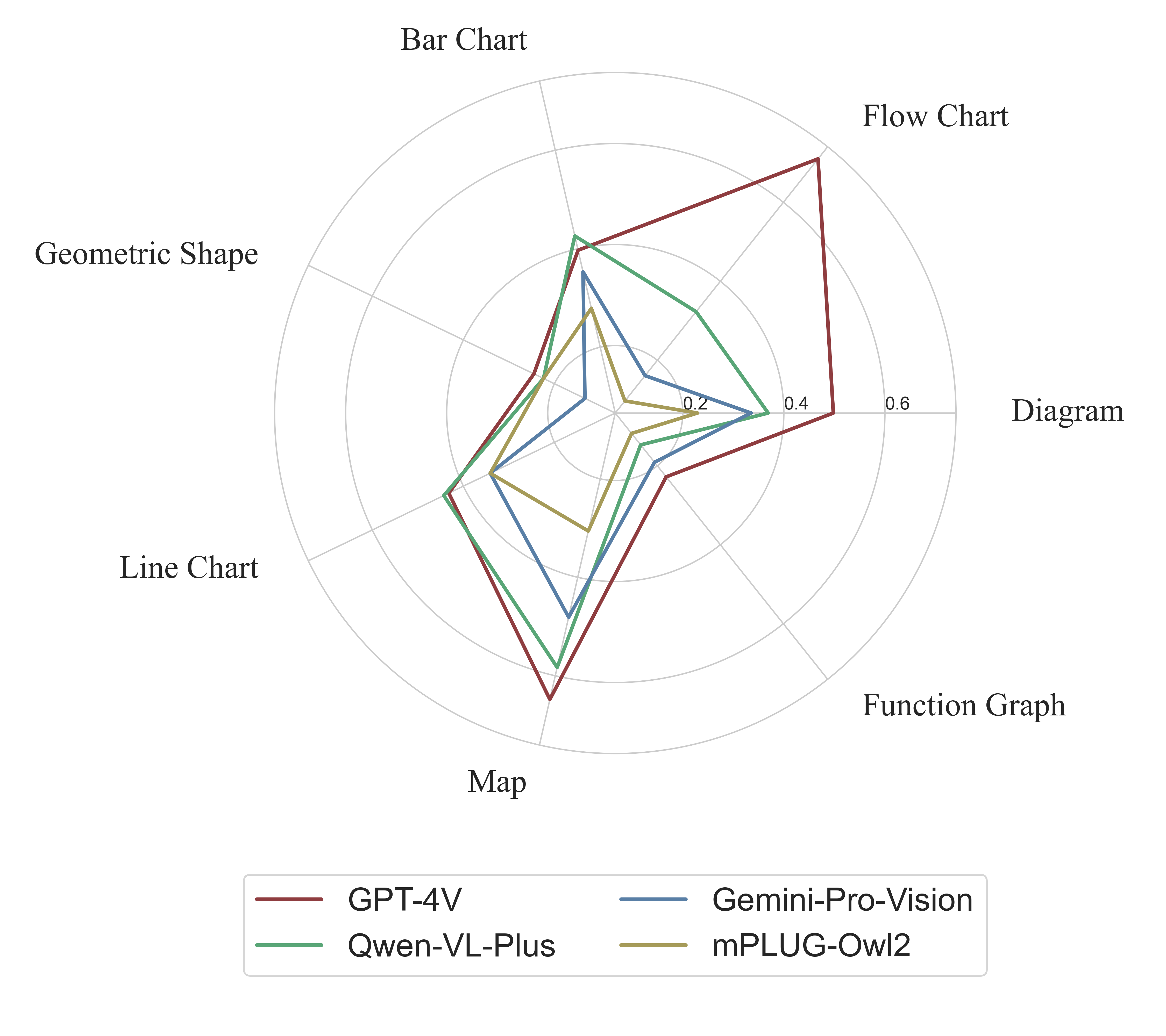}
    \end{adjustbox}
    \caption{Different Performance in Image Types.}
    \label{fig:difference in image types}
\end{figure}

\begin{figure*}[!htbp]
    \centering
    \begin{adjustbox}{center}
    \includegraphics[width=\linewidth]{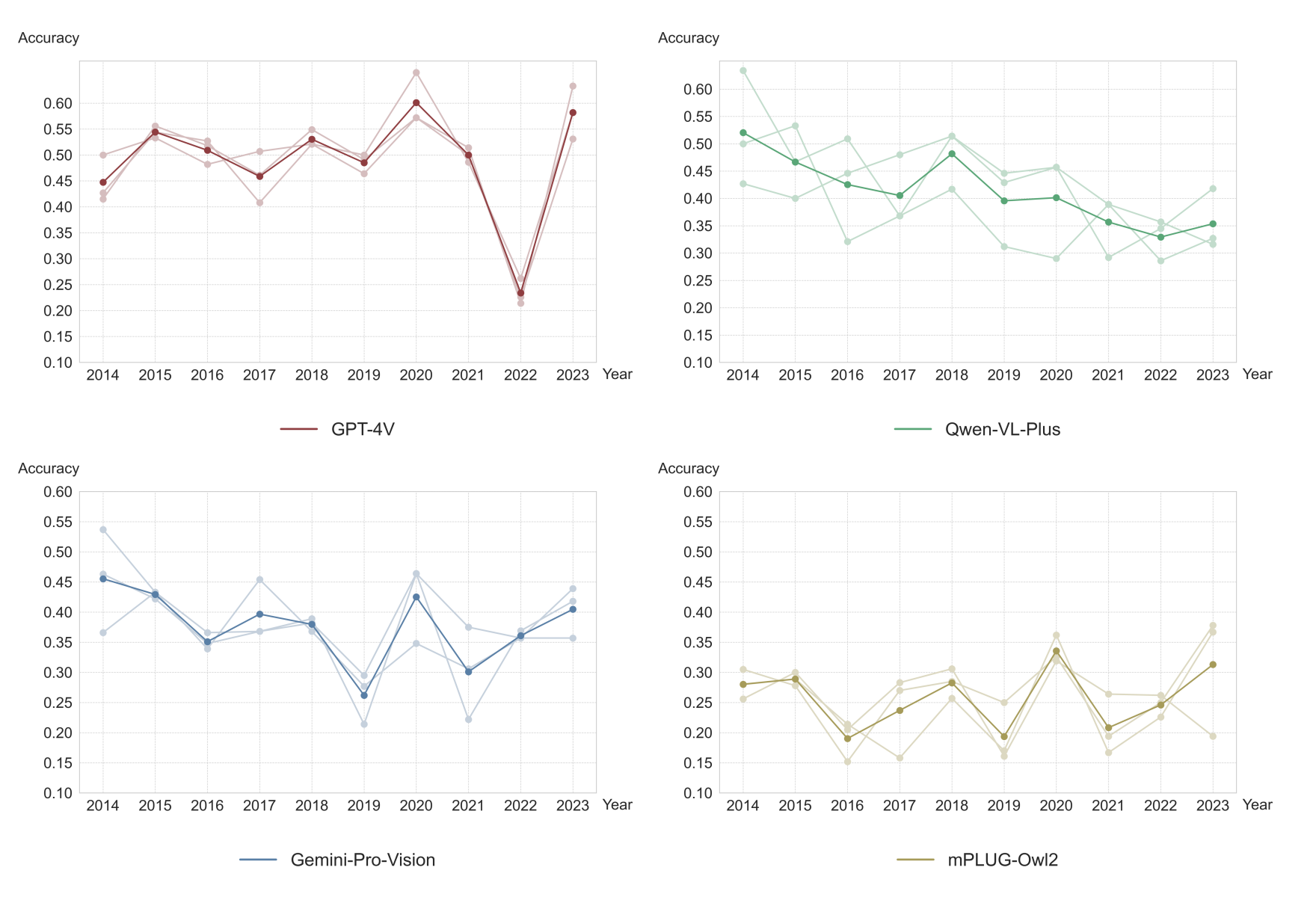}
    \end{adjustbox}
    \caption{\textbf{Difference in Annual Trends.}\quad The light-colored lines represent the accuracy obtained from three tests, while the dark-colored line represents the average accuracy. }
    \label{fig:difference in years}
\end{figure*}

\section{Analysis}
\subsection{Difference in Subjects}
Models perform differently among the 8 subjects as is shown in Figure \ref{fig:difference in subject}. Most models perform worse on mathematics and physics, which can be attributed to the the weak reasoning and calculation abilities. Notably, the accuracy of Chinese is the lowest for most models due to the reason that the average length of Chinese questions is the longest and models have to identify details in the text that correspond to the information in the images. It reveals that fine-grained understanding of long texts and images remains a significant challenge for LVLMs.




\subsection{Difference in Image Types}
We present the accuracy rates of the top 4 models on the primary image types in Figure \ref{fig:difference in image types}. All of them have accuracy rates exceeding their average accuracies on map-type images, while none of them excel in function graphs and geometric shapes. This is related to the imbalanced distribution of image types in image datasets and the varying demands of different images on specific domain knowledge. However, GPT-4V remarkably outperforms other models on flow charts, which reflects its superior logical reasoning abilities.

\subsection{Difference in Annual Trends}
We observe differences in the performance of models on questions from various years in Figure \ref{fig:difference in years}. The top four models do not exhibit consistent trends in performance over the past decade's questions. And the accuracy of the same model can vary by up to 20\% across different years. Considering the consistent difficulty and variety of questions in the GAOKAO each year, we believe that this fluctuation is related to the robustness of large-scale models. Their output quality is sensitive to the input questions and images. This might be one of the focal points for future research.


\section{Conclusion}
In this work, we propose GAOKAO-MM, a novel Chinese multimodal benchmark focusing on the human-level capabilities of LVLMs, including perception, understanding, knowledge and reasoning. 
Deriving from Chinese College Entrance Examination(GAOKAO) and comprising of diverse subjects and images, it is well-suited for accessing comprehensive abilities in Chinese context. 
We evaluate 10 prominent LVLMs such as GPT-4V, Gemini-Pro-Vision and Qwen-VL-Plus, and the accuracies of them are lower than 50\%, demonstrating considerable room for improvement. 
Our multi-dimension analysis reveal that the models' mathematical reasoning ability and comprehension of long texts and images still need to be enhanced, and the models' robustness to different texts and images is one of the future research directions. 
We believe GAOKAO-MM can propel the development of LVLMs towards AGI and provide insights into their applications in areas such as education.

\section*{Limitations}
While we make contributions to Chinese multimodal benchmarks, there are still some limitations in this work. 
Firstly, due to the cost of collecting high-quality data, our benchmark covers the GAOKAO questions from 2010 to 2023 and has room for improvement in terms of the scalability and balanced distribution of data. 
Secondly, the evaluation lacks detailed analysis of the inference process of LVLMs. We plan to maintain GAOKAO-MM as an ongoing project, supplementing it annually with new questions from the GAOKAO.

\section*{Ethics Statement}
In the process of developing GAOKAO-MM, we strictly adhere to ethical and legal standards. In China, the copyright of GAOKAO questions is generally considered to be in the public domain, allowing us to use them as an open evaluation dataset. We ensure that no private or sensitive information is included. We believe that open access to this dataset will contribute to the advancement of research in the field of multimodal models while maintaining high ethical standards.

\section*{Acknowledgement}
This work was supported by the National Key Research and Development Program of China (No.2022ZD0160102). The computations in this research were performed using the CFFF platform of Fudan University. 
We extend our gratitude to all reviewers for their valuable feedback, and we also wish to thank Botian Jiang, Bo Wang, Qingyuan Cheng and Jiachun Li for their insightful discussions.

\bibliography{custom}

\begin{thebibliography}{24}
\expandafter\ifx\csname natexlab\endcsname\relax\def\natexlab#1{#1}\fi

\bibitem[{Anil et~al.(2023)Anil, Dai, Firat, Johnson, and et~al.}]{anil2023palm}
Rohan Anil, Andrew~M. Dai, Orhan Firat, Melvin Johnson, and Dmitry~Lepikhin et~al. 2023.
\newblock \href {http://arxiv.org/abs/2305.10403} {Palm 2 technical report}.

\bibitem[{Bai et~al.(2023)Bai, Bai, Yang, Wang, Tan, Wang, Lin, Zhou, and Zhou}]{bai2023qwenvl}
Jinze Bai, Shuai Bai, Shusheng Yang, Shijie Wang, Sinan Tan, Peng Wang, Junyang Lin, Chang Zhou, and Jingren Zhou. 2023.
\newblock \href {http://arxiv.org/abs/2308.12966} {Qwen-vl: A versatile vision-language model for understanding, localization, text reading, and beyond}.

\bibitem[{Brown et~al.(2020)Brown, Mann, Ryder, Subbiah, Kaplan, Dhariwal, Neelakantan, Shyam, Sastry, Askell et~al.}]{brown2020language}
Tom Brown, Benjamin Mann, Nick Ryder, Melanie Subbiah, Jared~D Kaplan, Prafulla Dhariwal, Arvind Neelakantan, Pranav Shyam, Girish Sastry, Amanda Askell, et~al. 2020.
\newblock Language models are few-shot learners.
\newblock \emph{Advances in neural information processing systems}, 33:1877--1901.

\bibitem[{Bubeck et~al.(2023)Bubeck, Chandrasekaran, Eldan, Gehrke, Horvitz, Kamar, Lee, Lee, Li, Lundberg, Nori, Palangi, Ribeiro, and Zhang}]{bubeck2023sparks}
Sébastien Bubeck, Varun Chandrasekaran, Ronen Eldan, Johannes Gehrke, Eric Horvitz, Ece Kamar, Peter Lee, Yin~Tat Lee, Yuanzhi Li, Scott Lundberg, Harsha Nori, Hamid Palangi, Marco~Tulio Ribeiro, and Yi~Zhang. 2023.
\newblock \href {http://arxiv.org/abs/2303.12712} {Sparks of artificial general intelligence: Early experiments with gpt-4}.

\bibitem[{Du et~al.(2022)Du, Qian, Liu, Ding, Qiu, Yang, and Tang}]{du2022glm}
Zhengxiao Du, Yujie Qian, Xiao Liu, Ming Ding, Jiezhong Qiu, Zhilin Yang, and Jie Tang. 2022.
\newblock \href {http://arxiv.org/abs/2103.10360} {Glm: General language model pretraining with autoregressive blank infilling}.

\bibitem[{Fu et~al.(2023)Fu, Chen, Shen, Qin, Zhang, Lin, Yang, Zheng, Li, Sun, Wu, and Ji}]{fu2023mme}
Chaoyou Fu, Peixian Chen, Yunhang Shen, Yulei Qin, Mengdan Zhang, Xu~Lin, Jinrui Yang, Xiawu Zheng, Ke~Li, Xing Sun, Yunsheng Wu, and Rongrong Ji. 2023.
\newblock \href {http://arxiv.org/abs/2306.13394} {Mme: A comprehensive evaluation benchmark for multimodal large language models}.

\bibitem[{Goyal et~al.(2017)Goyal, Khot, Summers-Stay, Batra, and Parikh}]{goyal2017making}
Yash Goyal, Tejas Khot, Douglas Summers-Stay, Dhruv Batra, and Devi Parikh. 2017.
\newblock \href {http://arxiv.org/abs/1612.00837} {Making the v in vqa matter: Elevating the role of image understanding in visual question answering}.

\bibitem[{Hu et~al.(2024)Hu, Yao, Wang, Wang, Pan, Chen, Yu, Wu, Zhao, Zhang, Han, Lin, Xue, Li, Liu, and Sun}]{hu2024large}
Jinyi Hu, Yuan Yao, Chongyi Wang, Shan Wang, Yinxu Pan, Qianyu Chen, Tianyu Yu, Hanghao Wu, Yue Zhao, Haoye Zhang, Xu~Han, Yankai Lin, Jiao Xue, Dahai Li, Zhiyuan Liu, and Maosong Sun. 2024.
\newblock \href {http://arxiv.org/abs/2308.12038} {Large multilingual models pivot zero-shot multimodal learning across languages}.

\bibitem[{Kahou et~al.(2018)Kahou, Michalski, Atkinson, Kadar, Trischler, and Bengio}]{kahou2018figureqa}
Samira~Ebrahimi Kahou, Vincent Michalski, Adam Atkinson, Akos Kadar, Adam Trischler, and Yoshua Bengio. 2018.
\newblock \href {http://arxiv.org/abs/1710.07300} {Figureqa: An annotated figure dataset for visual reasoning}.

\bibitem[{Kembhavi et~al.(2016)Kembhavi, Salvato, Kolve, Seo, Hajishirzi, and Farhadi}]{kembhavi2016diagram}
Aniruddha Kembhavi, Mike Salvato, Eric Kolve, Minjoon Seo, Hannaneh Hajishirzi, and Ali Farhadi. 2016.
\newblock \href {http://arxiv.org/abs/1603.07396} {A diagram is worth a dozen images}.

\bibitem[{Liu et~al.(2023{\natexlab{a}})Liu, Li, Li, and Lee}]{liu2023improved}
Haotian Liu, Chunyuan Li, Yuheng Li, and Yong~Jae Lee. 2023{\natexlab{a}}.
\newblock \href {http://arxiv.org/abs/2310.03744} {Improved baselines with visual instruction tuning}.

\bibitem[{Liu et~al.(2023{\natexlab{b}})Liu, Duan, Zhang, Li, Zhang, Zhao, Yuan, Wang, He, Liu, Chen, and Lin}]{liu2023mmbench}
Yuan Liu, Haodong Duan, Yuanhan Zhang, Bo~Li, Songyang Zhang, Wangbo Zhao, Yike Yuan, Jiaqi Wang, Conghui He, Ziwei Liu, Kai Chen, and Dahua Lin. 2023{\natexlab{b}}.
\newblock \href {http://arxiv.org/abs/2307.06281} {Mmbench: Is your multi-modal model an all-around player?}

\bibitem[{Lu et~al.(2024)Lu, Bansal, Xia, Liu, Li, Hajishirzi, Cheng, Chang, Galley, and Gao}]{lu2024mathvista}
Pan Lu, Hritik Bansal, Tony Xia, Jiacheng Liu, Chunyuan Li, Hannaneh Hajishirzi, Hao Cheng, Kai-Wei Chang, Michel Galley, and Jianfeng Gao. 2024.
\newblock \href {http://arxiv.org/abs/2310.02255} {Mathvista: Evaluating mathematical reasoning of foundation models in visual contexts}.

\bibitem[{Lu et~al.(2022{\natexlab{a}})Lu, Mishra, Xia, Qiu, Chang, Zhu, Tafjord, Clark, and Kalyan}]{lu2022learn}
Pan Lu, Swaroop Mishra, Tony Xia, Liang Qiu, Kai-Wei Chang, Song-Chun Zhu, Oyvind Tafjord, Peter Clark, and Ashwin Kalyan. 2022{\natexlab{a}}.
\newblock \href {http://arxiv.org/abs/2209.09513} {Learn to explain: Multimodal reasoning via thought chains for science question answering}.

\bibitem[{Lu et~al.(2022{\natexlab{b}})Lu, Qiu, Chen, Xia, Zhao, Zhang, Yu, Liang, and Zhu}]{lu2022iconqa}
Pan Lu, Liang Qiu, Jiaqi Chen, Tony Xia, Yizhou Zhao, Wei Zhang, Zhou Yu, Xiaodan Liang, and Song-Chun Zhu. 2022{\natexlab{b}}.
\newblock \href {http://arxiv.org/abs/2110.13214} {Iconqa: A new benchmark for abstract diagram understanding and visual language reasoning}.

\bibitem[{Marino et~al.(2019)Marino, Rastegari, Farhadi, and Mottaghi}]{marino2019okvqa}
Kenneth Marino, Mohammad Rastegari, Ali Farhadi, and Roozbeh Mottaghi. 2019.
\newblock \href {http://arxiv.org/abs/1906.00067} {Ok-vqa: A visual question answering benchmark requiring external knowledge}.

\bibitem[{OpenAI(2023)}]{OpenAI2023GPT4TR}
OpenAI. 2023.
\newblock Gpt-4 technical report.
\newblock \emph{ArXiv}, abs/2303.08774.

\bibitem[{Singh et~al.(2019)Singh, Natarajan, Shah, Jiang, Chen, Batra, Parikh, and Rohrbach}]{singh2019vqa}
Amanpreet Singh, Vivek Natarajan, Meet Shah, Yu~Jiang, Xinlei Chen, Dhruv Batra, Devi Parikh, and Marcus Rohrbach. 2019.
\newblock \href {http://arxiv.org/abs/1904.08920} {Towards vqa models that can read}.

\bibitem[{Team et~al.(2023)Team, Anil, and et~al.}]{geminiteam2023gemini}
Gemini Team, Rohan Anil, and Sebastian~Borgeaud et~al. 2023.
\newblock \href {http://arxiv.org/abs/2312.11805} {Gemini: A family of highly capable multimodal models}.

\bibitem[{Wang et~al.(2024)Wang, Lv, Yu, Hong, Qi, Wang, Ji, Yang, Zhao, Song, Xu, Xu, Li, Dong, Ding, and Tang}]{wang2024cogvlm}
Weihan Wang, Qingsong Lv, Wenmeng Yu, Wenyi Hong, Ji~Qi, Yan Wang, Junhui Ji, Zhuoyi Yang, Lei Zhao, Xixuan Song, Jiazheng Xu, Bin Xu, Juanzi Li, Yuxiao Dong, Ming Ding, and Jie Tang. 2024.
\newblock \href {http://arxiv.org/abs/2311.03079} {Cogvlm: Visual expert for pretrained language models}.

\bibitem[{Ye et~al.(2023)Ye, Xu, Ye, Yan, Hu, Liu, Qian, Zhang, Huang, and Zhou}]{ye2023mplugowl2}
Qinghao Ye, Haiyang Xu, Jiabo Ye, Ming Yan, Anwen Hu, Haowei Liu, Qi~Qian, Ji~Zhang, Fei Huang, and Jingren Zhou. 2023.
\newblock \href {http://arxiv.org/abs/2311.04257} {mplug-owl2: Revolutionizing multi-modal large language model with modality collaboration}.

\bibitem[{Yu et~al.(2023)Yu, Yang, Li, Wang, Lin, Liu, Wang, and Wang}]{yu2023mmvet}
Weihao Yu, Zhengyuan Yang, Linjie Li, Jianfeng Wang, Kevin Lin, Zicheng Liu, Xinchao Wang, and Lijuan Wang. 2023.
\newblock \href {http://arxiv.org/abs/2308.02490} {Mm-vet: Evaluating large multimodal models for integrated capabilities}.

\bibitem[{Yue et~al.(2023)Yue, Ni, Zhang, Zheng, Liu, Zhang, Stevens, Jiang, Ren, Sun, Wei, Yu, Yuan, Sun, Yin, Zheng, Yang, Liu, Huang, Sun, Su, and Chen}]{yue2023mmmu}
Xiang Yue, Yuansheng Ni, Kai Zhang, Tianyu Zheng, Ruoqi Liu, Ge~Zhang, Samuel Stevens, Dongfu Jiang, Weiming Ren, Yuxuan Sun, Cong Wei, Botao Yu, Ruibin Yuan, Renliang Sun, Ming Yin, Boyuan Zheng, Zhenzhu Yang, Yibo Liu, Wenhao Huang, Huan Sun, Yu~Su, and Wenhu Chen. 2023.
\newblock \href {http://arxiv.org/abs/2311.16502} {Mmmu: A massive multi-discipline multimodal understanding and reasoning benchmark for expert agi}.

\bibitem[{Zhang et~al.(2023)Zhang, Li, Zong, Ying, He, and Qiu}]{zhang2023evaluating}
Xiaotian Zhang, Chunyang Li, Yi~Zong, Zhengyu Ying, Liang He, and Xipeng Qiu. 2023.
\newblock \href {http://arxiv.org/abs/2305.12474} {Evaluating the performance of large language models on gaokao benchmark}.

\end{thebibliography}

\begin{CJK*}{UTF8}{gbsn}

\appendix
\section{Key Statistics of GAOKAO-MM}
\label{app:statistics}
In this appendix, we show the distribution of image types in GAOKAO-MM in Figure \ref{fig:difference in image types}. The statistics of 8 subjects are shown in Table\ref{tab: statistics_q_t}. 

\begin{figure}[h]
    \centering
    \begin{adjustbox}{center}
    \includegraphics[width=1\linewidth]{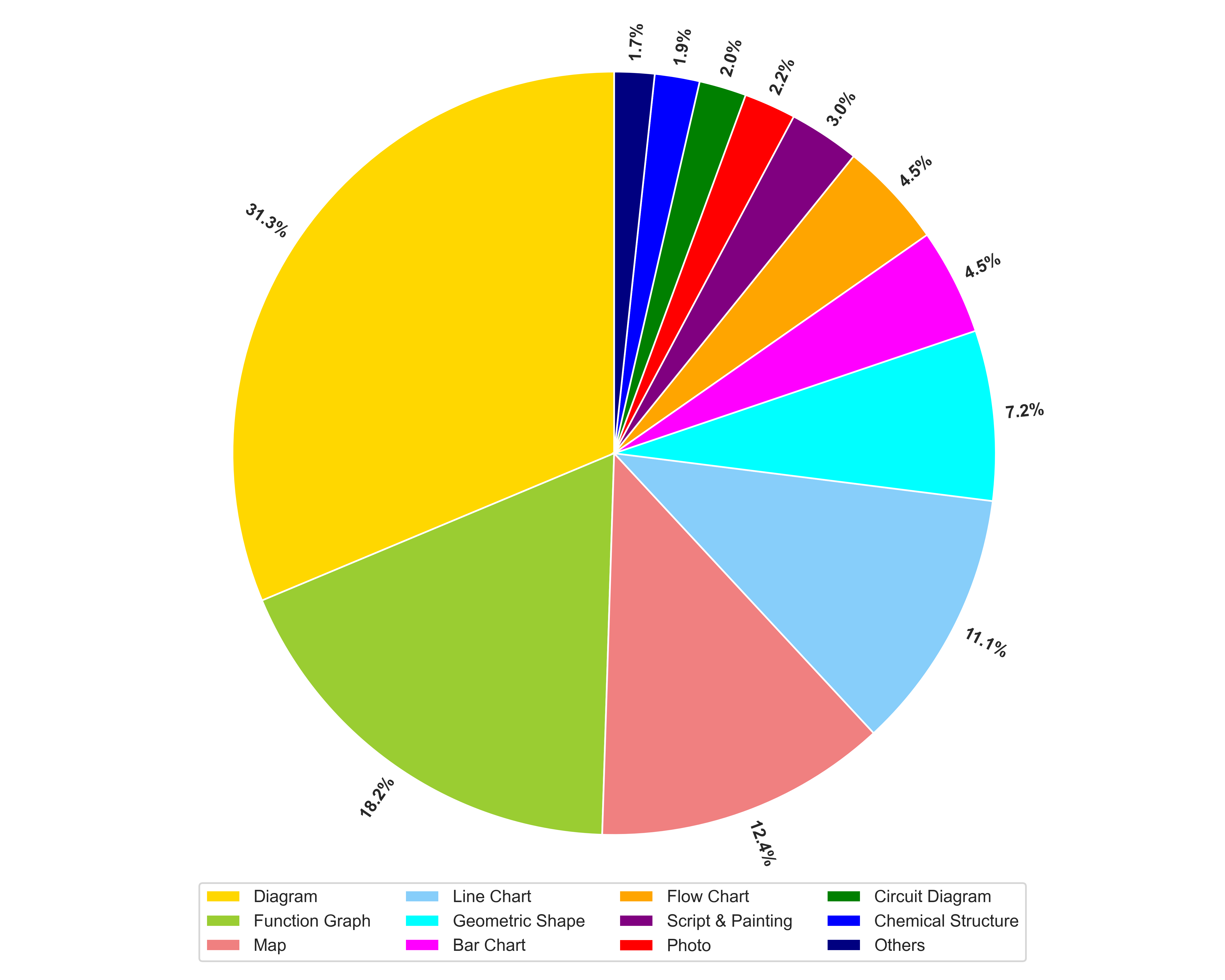}
    \end{adjustbox}
    \caption{Distribution of Image Types in GAOKAO-MM}
    \label{fig:image type distribution}
\end{figure}

\renewcommand{\arraystretch}{1.2}
\begin{table}[htb]
\centering
\resizebox{\linewidth}{!}{
\begin{tabular}{lccc}
\hline
\textbf{Question Type} & \textbf{Type} & \textbf{Number} & \textbf{Avg. Length}  \\
\hline
\textbf{Geography MCQs} &single choice & 221& 179.8 \\ 
\textbf{Physics MCQs} &multi choice & 174& 299.1 \\ 
\textbf{Math MCQs} &single choice & 80& 157.8 \\ 
\textbf{Chemistry MCQs} &single choice & 67& 399.1 \\ 
\textbf{History MCQs} &single choice & 34& 115.1 \\ 
\textbf{Political Science MCQs} &single choice & 33& 183.1 \\ 
\textbf{Biology MCQs} &single choice & 21& 257.0 \\ 
\textbf{Chinese Practical Lit} &multi choice & 16& 1349.3 \\ 
\hline
\end{tabular}
}
\caption{\label{tab: statistics_q_t}Statistics of Question Types. }
\end{table}
\renewcommand{\arraystretch}{1} 

\section{Examples}
In this appendix, we present examples of prompts used in the zero-shot setting experiment, questions and corresponding explanations and the outputs of GPT-4V, Qwen-VL-Plus, Gemini-Pro-Vision and mPLUG-Owl2. The English translation is added by the author for readers' understanding.

\subsection{Examples of Prompts}
\label{app:examples of prompts}
We design prompts to standardize the output format of the models and leverage their reasoning abilities. And we present an example of a single-choice question prompt and an example of a multiple-choice question prompt below.

\begin{tcolorbox}[breakable,colback=black!5!white,colframe=black!75!black]
\begin{itemize}
    \item \textbf{Single-Choice Questions}
\end{itemize}
请你做一道数学选择题。\\
请你结合文字和图片一步一步思考,并将思考过程写在【解析】和<eoe>之间。你将从A，B，C，D中选出正确的答案，并写在【答案】和<eoa>之间。\\
例如：【答案】: A <eoa>\\
完整的题目回答的格式如下：\\
【解析】 ...<eoe>\\
【答案】 ...<eoa>\\
请你严格按照上述格式作答。\\
题目如下：\\

Please solve the following mathematical multiple-choice question:\\
Please integrate textual explanations with accompanying diagrams to guide step-by-step problem-solving. Record your thought process between 【Analysis】and <eoe>. Select the correct answer from options A, B, C, and D, and indicate it between 【Answer】 and <eoa>.\\
For instance: 【Answer】A<eoa>\\
The format of the complete question answer is as follows:\\
【Analysis】 ...<eoe>\\
【Answer】 ...<eoa>\\
Ensure adherence to the prescribed format for comprehensive question-answering.\\
The question is as follows:\\

\begin{itemize}
    \item \textbf{Multiple-Choice Questions}
\end{itemize}
请你做一道语文阅读理解题。\\
请你结合文字和图片一步一步思考,并将思考过程写在【解析】和<eoe>之间。你将从A，B，C，D，E中选出所有符合题意的答案，并写在【答案】和<eoa>之间。\\
例如：【答案】 AB <eoa>\\
完整的题目回答的格式如下：\\
【解析】 ... <eoe>\\
【答案】... <eoa>\\
请你严格按照上述格式作答。\\
题目如下：\\

Please complete a Chinese reading comprehension question.
Please integrate textual explanations with accompanying diagrams to guide step-by-step problem-solving. Record your thought process between 【Analysis】and <eoe>. Select all suitable answers from A, B, C, D, and E, and indicate it between 【Answer】 and <eoa>.\\
For instance: 【Answer】AB <eoa>\\
The format of the complete question answer is as follows:\\
【Analysis】 ...<eoe> \\
【Answer】 ...<eoa>\\
Ensure adherence to the prescribed format for comprehensive question-answering.\\
The question is as follows:\\
\end{tcolorbox}



\subsection{Examples of Questions and Explanations}
\label{app:example of question and explanations}
In this appendix, we present an example of history questions and its explanations annotated by human.
Each question in GAOKAO-MM is equipped with detailed explanation to analysis the reasoning process of LVLMs. 

\begin{figure}[!htb]
    \centering
    \begin{adjustbox}{center}
    \includegraphics[width=\linewidth]{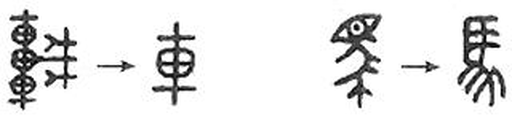}
    \end{adjustbox}
    \caption{Image of An Example of History Questions.}
    \label{fig:example of history quesitons}
\end{figure}

\begin{tcolorbox}[breakable,colback=black!5!white,colframe=black!75!black]
\begin{itemize}
    \item \textbf{Question}
\end{itemize}
1. 如图是西周与战国两个时期相同文字的不同写法, 反映出字形发生了变化, 促成这一变化的主要因素是 ( )\\\\
A. 文字的频繁使用	\\
B. 书写材料的不同	\\
C. 各国变法的实施	\\
D. “书同文”的推行 \\
\\

The picture shows the different writing methods of the same characters in the Western Zhou Dynasty and the Warring States Period, reflecting changes in the shape of the characters. The main factors that contributed to this change are ( )\\
A. Frequent use of writing	 		\\				
B. Differences in writing materials \\	
C. Implementation of reforms in various countries 	\\	  
D. Promotion of "shu tongwen"\\

\begin{itemize}
    \item \textbf{Answer}
\end{itemize}
A
\\
\begin{itemize}
    \item \textbf{Explanation}
\end{itemize}
【解答】\\
A. 中国古代汉字发展的总体趋势是由繁到简, 之所以出现这样的变化是由于中国的现实, 西周到战国时期社会政治经济市场化不断的发展, 文字在社会发展中的表述功能越来越强, 被频繁地使用, 所以必须由繁到简便于书写。\\
B. 从西周到战国时期, 我国出现了简读, 都是写在竹和木片上, 而且都用笔和墨。\\
C. 各国变法从春秋时期开始，但是各国变法中并没有涉及到文字的。\\
D. 书同文是秦始皇推行的措施, 和材料时代不符。\\
故选: A。\\
\\
【Analysis】\\
A. The overall trend of the development of ancient Chinese characters is from complexity to simplicity. The reason for such changes is due to the continuous development of society, politics, economy, and marketization from the Western Zhou Dynasty to the Warring States Period in China. As writing became increasingly important in social development, it was used frequently, hence the need for simplification for ease of writing. \\
B. From the Western Zhou Dynasty to the Warring States Period, simplification emerged in China, with writing done on bamboo and wooden slips, using pens and ink. \\
C. Reforms in various countries began in the Spring and Autumn Period, but they did not involve changes to writing systems. \\
D. "Shu tongwen" was a measure implemented by Emperor Qin Shi Huang, which is not consistent with the materials used during that period.\\
The answer is A.

\end{tcolorbox}


\subsection{Examples of Model Outputs}
\label{app:Examples of Model Outputs}
In this appendix, we show the corresponding outputs of the history questions mentioned above generated by GPT-4V, Qwen-VL-Plus, Gemini-Pro-Vision and mPLUG-Owl2. \\ 
The analysis provided by GPT-4V is the most comprehensive, but it fails to grasp the key aspect of the options; Qwen-VL-Plus and Gemini-Pro-Vision each analyzes only one option, with Qwen-VL-Plus overlooking crucial information in the question, while Gemini-Pro-Vision's reasoning and analysis are correct; mPLUG-Owl2, on the other hand, lacks a meaningful analytical process altogether.\\

\begin{tcolorbox}[breakable,colback=black!5!white,colframe=black!75!black]
\begin{itemize}
    \item \textbf{Qwen-VL-Plus Output}
\end{itemize}
【解析】
本题考查的是秦朝巩固统一的措施。秦始皇统一六国后，实行“书同文”，把七国文字统一成小篆，故选D。\\
【答案】 D \\
\\
This question examines the measures taken by the Qin Dynasty to consolidate its unification. After the unification of the six states by Emperor Qin Shi Huang, the policy of "shu tongwen" was implemented, unifying the scripts of the seven states into Xiaozhuan (small seal script). Therefore, option D is chosen.\\
Answer： D

\begin{itemize}
    \item \textbf{mPLUG-Owl2 Output}
\end{itemize}
【解析】: 
在这个问题中，我们需要找到西周和战国时期相同的文字，但是它们的写法不同。这表明字形发生了变化，因此我们需要找到这种变化的主要原因。\\
【答案】: B. 书写材料的不同\\
\\
In this question, we need to identify characters that are the same between the Western Zhou Dynasty and the Warring States Period but are written differently. This indicates changes in character shapes, so we need to identify the main reasons for these changes.\\
Answer： B. Differences in writing materials

\begin{itemize}
    \item \textbf{GPT-4V Output}
\end{itemize}
【解析】此题考查的是中国古代文字的发展变化。西周时期的文字主要是金文，而战国时期的文字则逐渐演变为秦篆。金文的书写多用于铸造在青铜器上，而秦篆的书写则更多地出现在竹简和帛书上。从西周到战国，书写材料的变化对文字的演变起到了重要作用。竹简和帛书的使用使得书写更加便捷，书写工具从刻刀变为毛笔，这促使了文字形态由繁复向简化发展。因此，书写材料的不同是促成字形变化的主要因素。选项A“文字的频繁使用”虽然也对文字的演变有一定影响，但不是主要因素；选项C“各国变法的实施”与文字演变的直接关系不大；选项D“‘书同文’的推行”是秦朝统一六国后的事情，与战国时期文字的变化无直接关系。因此，最合适的答案是B。 \\
【答案】: B \\
\\
This question examines the development and changes in ancient Chinese characters. During the Western Zhou Dynasty, the predominant script was the Jinwen (inscriptions on bronze vessels), while in the Warring States Period, the script gradually evolved into Qin Zhuan (seal script). Jinwen was mainly used for casting on bronze vessels, whereas Qin Zhuan appeared more frequently on bamboo slips and silk books. The change in writing materials from the Western Zhou Dynasty to the Warring States Period played a significant role in the evolution of characters. The use of bamboo slips and silk books made writing more convenient, and the transition from engraving tools to brushes facilitated the simplification of characters from complexity to simplicity. Therefore, the difference in writing materials is the primary factor contributing to changes in character shapes. Option A, "Frequent use of writing," although influencing the evolution of characters to some extent, is not the main factor. Option C, "Implementation of reforms in various countries," has little direct relevance to the evolution of characters. Option D, "Promotion of 'shu tongwen'," occurred after the unification of the six states by the Qin Dynasty and is not directly related to the changes in characters during the Warring States Period. Therefore, the most appropriate answer is B.\\
Answer： B

\begin{itemize}
    \item \textbf{Gemini-Pro-Vision Output}
\end{itemize}
【解析】从图中可以看出，西周时期的文字较为繁杂，而战国时期的文字较为简单。这一变化的主要原因是文字的频繁使用。随着社会的发展，人们对文字的需求量越来越大，为了提高书写的效率，就需要简化字形。此外，书写材料的不同也会导致字形的变化。西周时期，人们主要使用甲骨和金文来书写，而战国时期，人们开始使用竹简和纸张来书写。这些书写材料的特性不同，也对字形的变化产生了一定的影响。\\
【答案】A \\
\\
From the image, we can see that the characters during the Western Zhou Dynasty were more complex, while those during the Warring States Period were simpler. The main reason for this change is the frequent use of writing. With the development of society, there was an increasing demand for writing, leading to the need for simplifying character forms to enhance writing efficiency. Additionally, differences in writing materials can also lead to changes in character shapes. During the Western Zhou Dynasty, people primarily used oracle bones and Jinwen (inscriptions on bronze vessels) for writing, while during the Warring States Period, they began using bamboo slips and paper. The characteristics of these writing materials also had an impact on the changes in character shapes.

Answer： A
\end{tcolorbox}



\end{CJK*}

\end{document}